\documentclass[conference]{IEEEtran}

\usepackage{header}

\IEEEoverridecommandlockouts
\usepackage{cite}
\usepackage{amsmath,amssymb,amsfonts}
\usepackage{algorithmic}
\usepackage{graphicx}
\usepackage{textcomp}
\usepackage{xcolor}
\def\BibTeX{{\rm B\kern-.05em{\sc i\kern-.025em b}\kern-.08em
    T\kern-.1667em\lower.7ex\hbox{E}\kern-.125emX}}
    
\usepackage{multirow}
\usepackage{tabularx,ragged2e}
\usepackage{tabulary}
\usepackage{adjustbox}
\usepackage{arydshln}
\usepackage[hidelinks]{hyperref}
\usepackage{cleveref}
\usepackage[flushleft]{threeparttable}
\usepackage{url}
\usepackage{hhline}
\usepackage{multicol}
\usepackage{booktabs}
\usepackage{footmisc}
\usepackage{graphicx}
\usepackage{arydshln}
\usepackage{threeparttable}
\usepackage{soul,color}
\definecolor{lightgray}{rgb}{0.95,0.85,0.85}
\sethlcolor{lightgray}

\definecolor{darkblue}{rgb}{0, 0, 0.5}
\hypersetup{colorlinks=true, citecolor=darkblue, linkcolor=darkblue, urlcolor=darkblue}

\begin{document}

\title{Are Neural Language Models Good Plagiarists?\\A Benchmark for Neural Paraphrase Detection}


 \author{\IEEEauthorblockN{Jan Philip Wahle}
 \IEEEauthorblockA{\textit{University of Wuppertal}\\
 Wuppertal, Germany \\
 wahle@uni-wuppertal.de}
 \and
 \IEEEauthorblockN{Terry Ruas}
 \IEEEauthorblockA{\textit{University of Wuppertal}\\
 Wuppertal, Germany\\
 ruas@uni-wuppertal.de}
 \and
 \IEEEauthorblockN{Norman Meuschke}
 \IEEEauthorblockA{\textit{University of Wuppertal}\\
 Wuppertal, Germany \\
 meuschke@uni-wuppertal.de}
 \and
 \IEEEauthorblockN{Bela Gipp}
 \IEEEauthorblockA{\textit{University of Wuppertal}\\
 Wuppertal, Germany \\
 gipp@uni-wuppertal.de}
 }
\maketitle

\thispagestyle{firststyle}

\begin{abstract}
Neural language models such as BERT allow for human-like text paraphrasing. This ability threatens academic integrity, as it aggravates identifying machine-obfuscated plagiarism. We make two contributions to foster the research on detecting these novel machine-paraphrases. First, we provide the first large-scale dataset of documents paraphrased using the Transformer-based models  BERT, RoBERTa, and Longformer. The dataset includes paragraphs from scientific papers on arXiv, theses, and Wikipedia articles and their paraphrased counterparts (1.5M paragraphs in total). We show the paraphrased text maintains the semantics of the original source. Second, we benchmark how well neural classification models can distinguish the original and paraphrased text. The dataset and source code of our study are publicly available.
\end{abstract}


%
%
\section{Introduction}

Transformer-based language models~\cite{VaswaniSPU17} have reshaped natural language processing (NLP) and become the standard paradigm for most NLP downstream tasks~\cite{DevlinCLT19, BrownMRS20}. Now, these models are rapidly advancing to other domains such as computer vision~\cite{KhanNHZ21}. We anticipate Transformer-based models will similarly influence plagiarism detection research in the near future \cite{dehouche2021plagiarism}. Plagiarism is the use of ideas, concepts, words, or structures without proper source acknowledgment. Often plagiarists employ paraphrasing to conceal such practices~\cite{Foltynek2019}. Paraphrasing tools, such as \textit{SpinBot}\footnote{\label{fn_spinbot}\url{https://spinbot.com}} and \textit{SpinnerChief}\footnote{\url{https://spinnerchief.com/}}, facilitate the obfuscation of plagiarised content and threaten the effectiveness of plagiarism detection systems (PDS).

We expect that paraphrasing tools will abandon deterministic machine-paraphrasing approaches in favor of neural language models, which can incorporate intrinsic features from human language effectively~\cite{BrownMRS20}.
The ability of models such as GPT-3~\cite{BrownMRS20} to produce human-like texts raises major concerns in the plagiarism detection community as statistical and traditional machine learning solutions cannot distinguish semantically similar texts reliably~\cite{Foltynek2020}. Using Transformer-based models for the classification seems to be intuitive to counteract this new form of plagiarism. However, Transformer-based solutions typically require sufficiently large sets of labeled training data to achieve high classification effectiveness. As the use of neural language models for paraphrasing is a recent trend, data for the training of PDS is lacking.

This paper contributes to the development of future detection methods for paraphrased text by providing, to our knowledge, the first large-scale dataset of text paraphrased using Transformer-based language models. 
We study how word-embeddings and three Transformer-based models used for paraphrasing (BERT~\cite{DevlinCLT19}, RoBERTa~\cite{LiuOGD19}, and Longformer~\cite{BeltagyPC20}) perform in classifying paraphrased text to underline the difficulty of the task and the dataset's ability to reflect it.
The \textbf{dataset and source code} of our study are publicly available\footnote{\url{https://doi.org/10.5281/zenodo.4621403}\label{footnote-dataset}}. We grant access to the source code after accepting the terms and conditions designed to prevent misuse. Please see the repository for details.

\begin{table*}[!t]
    \footnotesize
	\centering
    \caption{\label{tab:dataset_examples} An illustrative sample for each paraphrasing model and data source. \hl{Red background} highlights changed tokens compared to the original version. The ellipsis ``..." indicates the remainder of the paragraph.}
	\begin{tabular}{p{14cm} c c}
		\toprule
        \textbf{Original Parapgraphs:}\\
        
        \multicolumn{3}{p{16cm}}{
            \begin{itemize}
                \item[--] A mathematically rigorous approach to quantum field theory based on operator algebras is called an algebraic quantum field theory...
                \item[--] "Nuts" contains 5 instrumental compositions written and produced by Streisand, with the exception of "The Bar", including additional writing from Richard Baskin. All of the songs were recorded throughout 1987...
                \item[--] Agriculture is the foundation for economic growth, development and poverty annihilation in developing countries. Ghana is endowed with a variety of mineral and agricultural product (Breisinger, 2008) Ghana is a country...
            \end{itemize}}\\
		\textbf{BERT Paraphrased} & \textbf{Source} & \textbf{MLM Prob.} \\
		\hline
        \hl{The} mathematically rigorous approach to quantum field theory based \hl{upon} operator \hl{equations} is called an algebraic \hl{Quantum} field theory... & arXiv & 0.15\\
		\\
		\textbf{RoBERTa Paraphrased} & & \\
		\hline
		"Nuts" contains \hl{five} instrumental compositions written or produced by Streisand, with the exception of \hl{"Yourbars"}, \hl{which includes credited} writing from Richard Baskin. All of these songs were recorded \hl{in} 1987... & Wikipedia & 0.15\\
		\\
		\textbf{Longformer Paraphrased} & & \\
		\hline
		Agriculture is the foundation \hl{builder for} economic growth, development and poverty annihilation in developing countries. Ghana is endowed with a variety of \hl{biodiversity} and agricultural product (Breisinger, 2008) Ghana \hl{became} a country... & thesis & 0.15 \\
        
		\bottomrule
	\end{tabular}
\end{table*}

%
%
\section{Related Work}
Paraphrase identification is a well-researched NLP problem with numerous applications, e.g., in information retrieval and digital library research~\cite{Foltynek2019}. To identify paraphrases, many approaches combine lexical, syntactical, and semantic text analysis~\cite{Foltynek2020}. The Microsoft Research Paraphrase Corpus (MRPC)~\cite{DolanB05}, a collection of human-annotated sentence pairs extracted from news articles, is among the most widely-used datasets for training and evaluating paraphrase identification methods. Another popular resource for paraphrase identification is the \textit{Quora Question Pairs} (QQP) dataset included in GLUE~\cite{WangSMH19}. The dataset consists of questions posted on Quora\footnote{\url{https://quora.com/about}}, a platform on which users can ask for answers to arbitrary questions. The task is to identify questions in the dataset that share the same intent. The datasets published as part of the PAN workshop series on plagiarism detection, authorship analysis, and other forensic text analysis tasks\footnote{\url{https://pan.webis.de/}} are the most comprehensive and widely-used resource for evaluating plagiarism detection systems. 

Neither the PAN nor the MRPC and QQP datasets include paraphrases created using state-of-the-art neural language models. The MRPC and QQP datasets consist of human-made content, which is unsuitable for training classifiers to recognize machine-paraphrased text. 
The PAN datasets contain cases that were obfuscated using basic automated heuristics that do not maintain the meaning of the text.
Examples of such heuristics include randomly removing, inserting, or replacing words or phrases and substituting words with their synonyms, antonyms, hyponyms, or hypernyms selected at random~\cite{PotthastSBR10}. These cases are not representative of the sophisticated paraphrases produced by state-of-the-art Transformer-based models. 

Currently, the HuggingFace API offers few neural language models capable of paraphrasing text excerpts. Most models are based on the same technique and trained to process short sentences. Plagiarists reuse paragraphs most frequently~\cite{Foltynek2019}. Hence, the ability to identify paragraph-sized paraphrases is most relevant for a PDS in practice. Prior to our study, no dataset of paragraphs paraphrased using Transformer-based models existed and could be used for training PDS.

Prior studies mitigated the lack of suitable datasets by paraphrasing documents using the paid services SpinBot and SpinnerChief~\cite{Foltynek2020,Wahle21}. As the evaluations in these studies showed, text obfuscated by these tools already poses a significant challenge to current plagiarism detection systems. Nevertheless, the sophistication of the paraphrased text obtained from such tools to date is lower than that of paraphrases generated by Transformer-based models. Therefore, we extend the earlier studies~\cite{Foltynek2020} and \cite{Wahle21} by using Transformer-based architectures~\cite{VaswaniSPU17} to generate paraphrases that reflect the strongest level of disguise technically feasible to date. 



%
%
\section{Dataset Creation}\label{sec:dataset_creation}

Our neural machine-paraphrased dataset is derived from previous studies \cite{Foltynek2020, Wahle21}. The dataset of Foltynek et al. \cite{Foltynek2020} consists of \textit{featured Wikipedia articles}\footnote{\url{https://en.wikipedia.org/wiki/Wikipedia:Content_assessment}} in English. The dataset of Wahle et al. \cite{Wahle21} comprises scientific papers randomly sampled from the \textit{no problems} category of the arXMLiv\footnote{\url{https://kwarc.info/projects/arXMLiv/}} project, and randomly selected graduation theses by \textit{English as a Second Language} (ESL) students at the Mendel University in Brno, Czech Republic.

The earlier studies employed the paid online paraphrasing services SpinBot and SpinnerChief for text obfuscation. Since we investigate neural language models for paraphrasing, we only use the 163\,715 original paragraphs from the earlier dataset. Table~\ref{tab:dataset-description} shows the composition of these original paragraphs used for our dataset. 

\begin{table}[t!]
\caption{Overview of the original paragraphs in our dataset.\label{tab:dataset-description}}
\centering
\begin{tabular}{clrrrr}
\toprule
\textbf{Features} & \textbf{arXiv}    & \textbf{Theses}   & \textbf{Wiki} & \textbf{Wiki-Train} \\ \midrule
Paragraphs & 20\,966    & 5\,226     & 39\,241     & 98\,282       \\
\# Words   & 3\,194\,695  & 747\,545   & 5\;993\,461   & 17\,390\,048    \\
Avg. Words & 152.38   & 143.04   & 152.73    & 176.94      \\ \bottomrule   
\end{tabular}
\end{table}

For paraphrasing, we used BERT~\cite{DevlinCLT19}, RoBERTa~\cite{LiuOGD19}, and Longformer~\cite{BeltagyPC20}. We chose BERT as a strong baseline for transformer-based language models; RoBERTa and Longformer improve BERT's architecture through more training volume and an efficient attention mechanism, respectively. More specifically, we used the masked language model (MLM) objective of all three Transformer-based models to create the paraphrases. The MLM hides a configurable portion of the words in the input, for which the model then has to infer the most probable word-choices. We excluded named entities and punctuation, e.g., brackets, digits, currency symbols, quotation marks from paraphrasing to avoid producing false information, or inconsistent punctuation compared to the original source. Then, we masked words and forwarded them to each model to obtain word candidates and their confidence scores. Lastly, we replaced each masked word in the original with the corresponding candidate word having the highest confidence score. Examples of original and paraphrased text using different models and data sources are illustrated in \Cref{tab:dataset_examples}.
We also experimented with sampling uniformly from the top-k word predictions but neglected this method because of poor paraphrasing quality.


We ran an ablation study to understand how the masking probability of the MLM affects the difficulty of classifying documents as either paraphrased or original. For this purpose, we employed each neural language model with varying masking probabilities to paraphrase the arXiv, theses, and Wikipedia subsets. We encoded all original and paraphrased texts as features using the sentence embedding of fastText (subword)\footnote{\url{https://fasttext.cc/docs/en/english-vectors.html}}, which was trained on a 2017 dump of the full English Wikipedia, the UMBC WebBase corpus, and StatMT news dataset with 300 dimensions. Lastly, we applied the same SVM classifier to all fastText feature representations to distinguish between original and paraphrased content.

\begin{figure}[t!]
    \centering
    \includegraphics[scale=0.4]{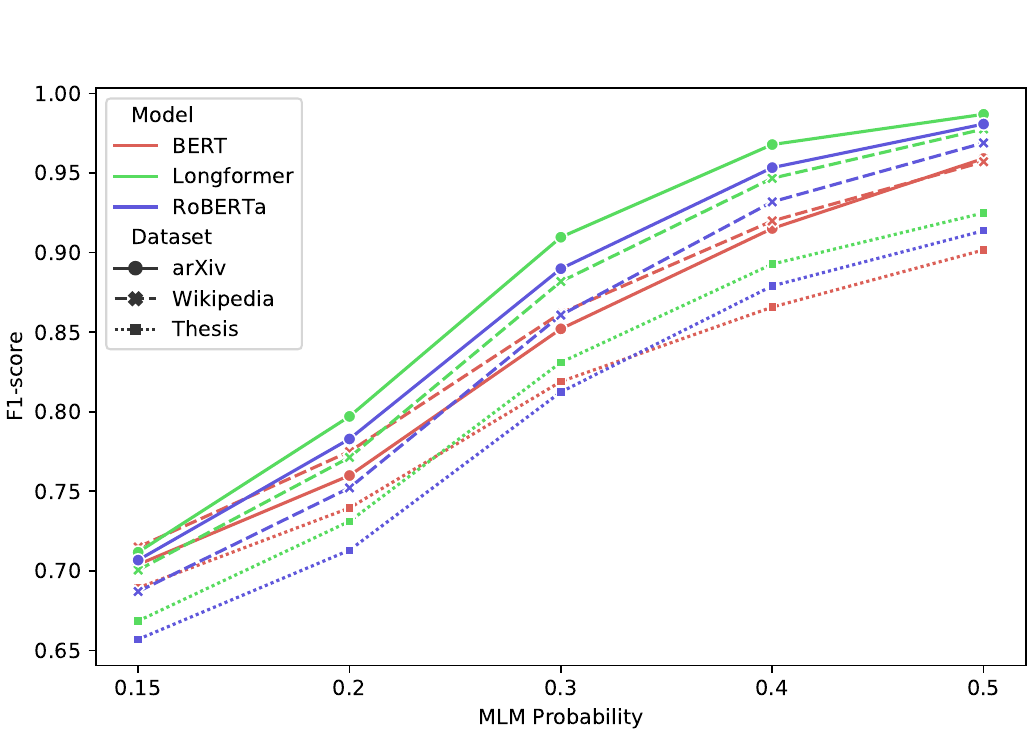}
    \caption{Classification accuracy of fastText + SVM for neural-paraphrased test sets depending on masked language model probabilities.}
    \label{fig:mlm_exp}
\end{figure}

Fig.~\ref{fig:mlm_exp} shows the results of the ablation study. Higher masking probabilities consistently led to higher classification accuracy. In other terms, replacing more words reduced the difficulty of the classification task. This correlation has also been observed for non-neural paraphrasing tools~\cite{Wahle21}. Paragraphs from theses were most challenging for the classifier regardless of the paraphrasing model. We hypothesize that sub-optimal word choice and grammatical errors in the texts written by ESL students increase the difficulty of classifying these texts. The F1-scores for paragraphs from arXiv and Wikipedia were consistently higher than for theses. We attribute the high score on the Wikipedia test set to the documents' similarity with the training set which consists only of Wikipedia articles.

Masking 15\% of the words posed the hardest challenge for the classifier. This ratio corresponds to the masking probability used for pre-training BERT~\cite{DevlinCLT19}, and falls into the percentage range of words that paid online paraphrasing tools replace on average (12.58\% to 19.37\%)~\cite{Wahle21}. Thus, we used a masking probability of 15\% for creating all paraphrased data. 

\begin{figure}[t!]
    \centering
    \includegraphics[scale=0.418]{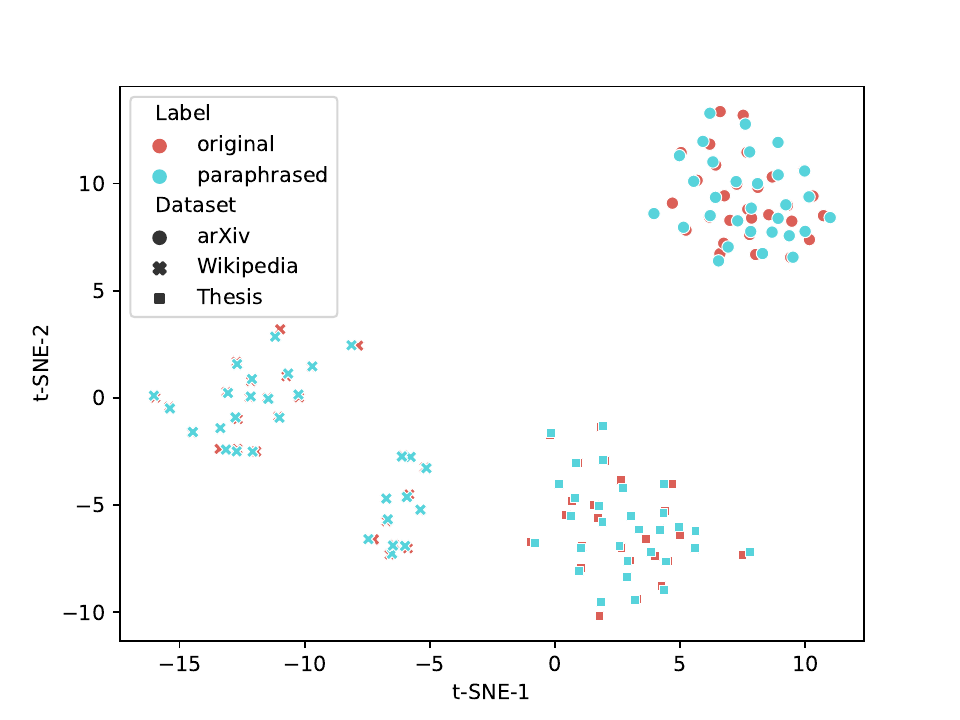}
    \caption{Two-dimensional representation of BERT embeddings for 30 original and paraphrased paragraphs from each source. The overlap of the embeddings suggests semantic equivalence of the original and paraphrased content.}
    \label{fig:paragraph_embedding}
\end{figure}

As a proxy for paraphrasing quality, we evaluated the semantic equivalence of original and paraphrased text. Specifically, we analyzed the BERT embeddings of 30 randomly selected original paragraphs from arXiv, theses, and Wikipedia and their paraphrased counterparts created using BERT, RoBERTa, and Longformer. Fig.~\ref{fig:paragraph_embedding} visualizes the results using a t-distributed Stochastic Neighbor Embedding (t-SNE) for dimensionality reduction. The embeddings of original and paraphrased text overlap considerably despite changing approx. 15\% of the words. This indicates the Transformer-based language models maintain the original text's semantics. 

%
%
\section{Classification Benchmark}\label{sec:benchmark}

To check whether our dataset poses a realistic challenge for state-of-the-art classifiers and to establish a performance benchmark, we employed four models to label paragraphs as either original or paraphrased. A prior study showed that current plagiarism detection systems, which are essentially text-matching software, fail to identify machine-paraphrased text reliably while word embeddings, machine-learning classifiers, and particularly Transformer-based models performed considerably better~\cite{Wahle21}. Therefore, we evaluated the classification effectiveness of the three BERT-related models used for paraphrasing and the fastText + SVM classifier we applied and described for our ablation study (cf. \Cref{sec:dataset_creation}). We limited the number of input tokens for each model to 512 for a fair comparison of the models without losing relevant context information\footnote{99.35\% of the datasets' text can be represented with less than 512 tokens.}. Unless specified differently, we used all hyperparameters in their default configuration.

We derived training data exclusively from Wikipedia as it is the largest of the three collections. We used arXiv papers and theses to obtain test sets that allow verifying a model's ability to generalize to data from sources unseen during training. We used BERT to generate the paraphrased training set (Wiki-Train) and BERT, RoBERTa, and Longfomer to create three paraphrased test sets. The classification models were exposed to mutually exclusive paragraphs to avoid memorizing the differences between aligned paragraphs. Evaluating each model using text paraphrased by the same model allows us to verify an assumption from related work, i.e., the best classifier is the language model used to generate the paraphrased text\cite{ZellersHRB19}.

\begin{table}[ht!]
\caption{Classification results (F1-Macro scores). \textbf{Boldface} shows the best result per classification model.}\label{tab:classification-results}
\resizebox{1\columnwidth}{!}{
\centering
\begin{tabular}{clrrr}
\toprule
\multirow{2}{*}{\textbf{Classification Model}} & \multirow{2}{*}{\textbf{Dataset}} & \multicolumn{3}{c}{\textbf{Paraphrase Model}} \\ \cmidrule(lr){3-5}

{} & {} & \textbf{BERT}    & \textbf{RoBERTa}   & \textbf{Longformer} \\ \midrule
\multirow{4}{*}{\shortstack[c]{fastText + SVM \\ (baseline)}}   & arXiv   & 70.40\%  & 70.68\%  & 71.17\%  \\
                            & Theses     & 68.94\%  & 65.70\%  & 66.85\%  \\ 
                             & Wikipedia       & 71.50\%  & 68.70\%  & 70.05\% \\ \cmidrule(lr){2-5}
                            & Average    & 70.28\%   & 68.36\%   & 69.36\%   \\ \midrule
\multirow{4}{*}{BERT}       & arXiv    & 80.83\%  & 68.90\%  & 68.49\%    \\
                            & Theses     & 74.74\%  & 67.39\%  & 66.04\%     \\ 
                            & Wikipedia       & 83.21\%  & 68.85\%  & 69.46\%    \\ \cmidrule(lr){2-5}
                            & Average    & \textbf{79.59\%}   & 68.38\%   & 68.00\% \\ \midrule
\multirow{4}{*}{RoBERTa}    & arXiv   & 70.41\%   & 85.40\%   & 82.95\%     \\
                            & Theses     & 68.99\%   & 79.13\%   & 77.76\%      \\ 
                            & Wikipedia        & 72.18\%   & 84.20\%   & 82.15\%     \\ \cmidrule(lr){2-5}
                            & Average    & 70.53\%   & \textbf{82.91\%}   & 80.95\%    \\ \midrule
\multirow{4}{*}{Longformer} & arXiv   & 65.18\%   & 85.46\%   & 89.93\%     \\
                            & Theses          & 65.72\%   & 77.96\%   & 81.31\%      \\ 
                            & Wikipedia       & 69.98\%   & 81.76\%   & 86.03\%      \\ \cmidrule(lr){2-5}
                            & Average         & 66.96\%   & 81.73\%   & \textbf{85.76}\%     \\ \bottomrule
\end{tabular}
}
\end{table} 

Table~\ref{tab:classification-results} shows the F1-Macro scores of each classification model for the paraphrased test sets consisting of arXiv, theses, and Wikipedia paragraphs. The baseline model (fastText + SVM) performed similarly for all paraphrasing models with scores ranging from F1=68.36\% (RoBERTa) to F1=70.28\% (BERT). With scores ranging from F1=79.59\% (BERT) to F1=85.76\% (Longformer), neural language models consistently identified text paraphrased using the same model best. This observation supports the findings of Zellers et al.~\cite{ZellersHRB19}.

Neural language models applied to paraphrases created by other models (e.g., BERT classifies text paraphrased by Longformer), typically achieved comparable scores to fastText+SVM. The average F1-scores for text paraphrased by unseen models range from F1=68.00\% (BERT for Longformer paraphrases) to F1=81.73\% (Longformer for RoBERTa paraphrases) with an average of 72.75\%. These results are lower than the average scores for classifying paraphrases created for the same subsets using paid paraphrasing services (i.e., F1=99.65\% to F1=99.87\% for SpinBot)~\cite{Wahle21}. This finding shows Transformer-based neural language models produce hard-to-identify paraphrases, which make our new dataset a challenging benchmark task for state-of-the-art classifiers.

RoBERTa and Longformer achieved comparable results for all datasets, which we attribute to their overlapping pre-training datasets. BERT uses a subset of RoBERTa's and Longformer's training data and identifies the text paraphrased by the other two models with comparable F1-scores.
Averaged over all paraphrasing techniques, RoBERTa achieved the best result (F1=78.15\%), making it the most general model we tested for detecting neural machine-paraphrases. 

All classification models performed best for Wikipedia articles, which is expected given their overlapping training corpus. The three neural language models identified arXiv articles similarly well which is in line with our ablation study (cf. Fig.~\ref{fig:mlm_exp}). As in our ablation study, theses by ESL students were most challenging for our classification models, again corroborating our assumption that a higher ratio of grammatical and linguistic errors causes the drop in classification effectiveness.

%
%
\section{Conclusion and Future Work}
We presented a large-scale aligned dataset\footref{footnote-dataset} of original and machine-paraphrased paragraphs to foster the research on plagiarism detection methods. The paragraphs originate from arXiv papers, theses, and Wikipedia articles and have been paraphrased using BERT, RoBERTa, and Longformer. 
We showed that the machine-paraphrased texts have a high semantic similarity to their original sources which reinforces our manual observation that neural language models produce hard-to distinguish, human-like paraphrases. 

Furthermore, we showed Transformers are comparable in classifying original and paraphrased content to static word embeddings (i.e., fastText) and most effective for identifying text that was paraphrased using the same model. RoBERTa achieved the best overall result for detecting paraphrases.

In our future work, we will investigate other autoencoding models, and add autoregressive models to our study such as GPT-3~\cite{BrownMRS20} for paraphrase generation and detection.

%
%


\bibliographystyle{IEEEtran}
\bibliography{IEEEabrv,_paper}

\begin{thebibliography}{10}
\providecommand{\url}[1]{#1}
\csname url@samestyle\endcsname
\providecommand{\newblock}{\relax}
\providecommand{\bibinfo}[2]{#2}
\providecommand{\BIBentrySTDinterwordspacing}{\spaceskip=0pt\relax}
\providecommand{\BIBentryALTinterwordstretchfactor}{4}
\providecommand{\BIBentryALTinterwordspacing}{\spaceskip=\fontdimen2\font plus
\BIBentryALTinterwordstretchfactor\fontdimen3\font minus
  \fontdimen4\font\relax}
\providecommand{\BIBforeignlanguage}[2]{{%
\expandafter\ifx\csname l@#1\endcsname\relax
\typeout{** WARNING: IEEEtran.bst: No hyphenation pattern has been}%
\typeout{** loaded for the language `#1'. Using the pattern for}%
\typeout{** the default language instead.}%
\else
\language=\csname l@#1\endcsname
\fi
#2}}
\providecommand{\BIBdecl}{\relax}
\BIBdecl

\bibitem{VaswaniSPU17}
A.~Vaswani, N.~Shazeer, N.~Parmar, J.~Uszkoreit, L.~Jones, A.~N. Gomez,
  L.~Kaiser, and I.~Polosukhin, ``Attention is all you need,'' in
  \emph{Advances in neural information processing systems 30}.\hskip 1em plus
  0.5em minus 0.4em\relax Curran Associates, Inc., 2017, pp. 5998--6008.

\bibitem{DevlinCLT19}
J.~Devlin, M.-W. Chang, K.~Lee, and K.~Toutanova, ``{BERT}: {Pre}-training of
  {Deep} {Bidirectional} {Transformers} for {Language} {Understanding},''
  \emph{arXiv:1810.04805 [cs]}, May 2019, arXiv: 1810.04805.

\bibitem{BrownMRS20}
T.~B. Brown, B.~Mann, N.~Ryder, M.~Subbiah, and D.~...~Amodei, ``Language
  {Models} are {Few}-{Shot} {Learners},'' \emph{arXiv:2005.14165 [cs]}, Jun.
  2020, tex.ids: BrownMRS20a arXiv: 2005.14165.

\bibitem{KhanNHZ21}
S.~Khan, M.~Naseer, M.~Hayat, S.~W. Zamir, F.~S. Khan, and M.~Shah,
  ``Transformers in {Vision}: {A} {Survey},'' \emph{arXiv:2101.01169 [cs]},
  Feb. 2021, arXiv: 2101.01169.

\bibitem{dehouche2021plagiarism}
N.~Dehouche, ``Plagiarism in the age of massive generative pre-trained
  transformers (gpt-3),'' \emph{Ethics in Science and Environmental Politics},
  vol.~21, pp. 17--23, 2021.

\bibitem{Foltynek2019}
T.~Folt{\'y}nek, N.~Meuschke, and B.~Gipp, ``Academic {Plagiarism} {Detection}:
  {A} {Systematic} {Literature} {Review},'' \emph{ACM Computing Surveys},
  vol.~52, no.~6, pp. 112:1--112:42, 2019.

\bibitem{Foltynek2020}
T.~Folt{\'y}nek, T.~Ruas, P.~Scharpf, N.~Meuschke, M.~Schubotz, W.~Grosky, and
  B.~Gipp, ``Detecting {Machine}-obfuscated {Plagiarism},'' in
  \emph{Proceedings of the {iConference} 2020}, ser. LNCS.\hskip 1em plus 0.5em
  minus 0.4em\relax Springer, 2020.

\bibitem{LiuOGD19}
Y.~Liu, M.~Ott, N.~Goyal, J.~Du, M.~Joshi, D.~Chen, O.~Levy, M.~Lewis,
  L.~Zettlemoyer, and V.~Stoyanov, ``{RoBERTa}: {A} {Robustly} {Optimized}
  {BERT} {Pretraining} {Approach},'' \emph{arXiv:1907.11692 [cs]}, Jul. 2019,
  arXiv: 1907.11692.

\bibitem{BeltagyPC20}
I.~Beltagy, M.~E. Peters, and A.~Cohan, ``Longformer: {The} {Long}-{Document}
  {Transformer},'' \emph{arXiv:2004.05150 [cs]}, Apr. 2020, arXiv: 2004.05150.

\bibitem{DolanB05}
B.~Dolan and C.~Brockett, ``Automatically constructing a corpus of sentential
  paraphrases,'' in \emph{Third International Workshop on Paraphrasing
  (IWP2005)}.\hskip 1em plus 0.5em minus 0.4em\relax Asia Fed. of Natural
  Language Processing, January 2005.

\bibitem{WangSMH19}
A.~Wang, A.~Singh, J.~Michael, F.~Hill, O.~Levy, and S.~R. Bowman, ``{{GLUE}}:
  {{A Multi}}-{{Task Benchmark}} and {{Analysis Platform}} for {{Natural
  Language Understanding}},'' \emph{arXiv:1804.07461 [cs]}, Feb. 2019.

\bibitem{PotthastSBR10}
M.~Potthast, B.~Stein, A.~Barr\'{o}n-Cede\~{n}o, and P.~Rosso, ``An
  {{Evaluation Framework}} for {{Plagiarism Detection}},'' in \emph{Proceedings
  {{Int. Conf.}} on {{Computational Linguistics}}}, vol.~2, pp. 997--1005.

\bibitem{Wahle21}
J.~P. Wahle, T.~Ruas, T.~Folt{\'y}nek, N.~Meuschke, and B.~Gipp, ``Identifying
  {{Machine}}-{{Paraphrased Plagiarism}},'' \emph{arXiv:2103.11909 [cs]}, Jan.
  2021.

\bibitem{ZellersHRB19}
R.~Zellers, A.~Holtzman, H.~Rashkin, Y.~Bisk, A.~Farhadi, F.~Roesner, and
  Y.~Choi, ``Defending {Against} {Neural} {Fake} {News},''
  \emph{arXiv:1905.12616 [cs]}, Oct. 2019, arXiv: 1905.12616.

\end{thebibliography}
\end{document}